\documentclass[10pt,twocolumn,letterpaper]{article}

\usepackage{cvpr}
\usepackage{times}
\usepackage{epsfig}
\usepackage{graphicx}
\usepackage{amsmath}
\usepackage{amssymb}
\usepackage{subfigure}

\usepackage[pagebackref=true,breaklinks=true,letterpaper=true,colorlinks,bookmarks=false]{hyperref}
\cvprfinalcopy % *** Uncomment this line for the final submission

\def\cvprPaperID{3546} % *** Enter the CVPR Paper ID here

\ifcvprfinal\fi
\begin{document}
\newcommand{\tabincell}[2]{\begin{tabular}{@{}#1@{}}#2\end{tabular}}
%%%%%%%%% TITLE
\newcommand{\argmax}[1]{\underset{#1}{\operatorname{arg}\,\operatorname{max}}\;}

\title{Action Machine: Rethinking Action Recognition in Trimmed Videos}

%\author{First Author\\
%Institution1\\
%Institution1 address\\
%{\tt\small firstauthor@i1.org}
%% For a paper whose authors are all at the same institution,
%% omit the following lines up until the closing ``}''.
%% Additional authors and addresses can be added with ``\and'',
%% just like the second author.
%% To save space, use either the email address or home page, not both
%\and
%Second Author\\
%Institution2\\
%First line of institution2 address\\
%{\tt\small secondauthor@i2.org}
%}

\author{Jiagang Zhu$^{1,2}$, Wei Zou$^{1}$, Liang Xu$^{3}$, Yiming Hu$^{1,2}$, Zheng Zhu$^{1,2}$, Manyu Chang$^{4}$,\\ Junjie Huang$^{1,2}$, Guan Huang$^{3}$, Dalong Du$^{3}$\\
$^{1}$Institute of Automation, Chinese Academy of Sciences (CASIA)\\
$^{2}$University of Chinese Academy of Sciences (UCAS)\\
$^{3}$Horizon Robotics, Inc.
$^{4}$Xiamen University\\
{\tt\small \{zhujiagang2015, wei.zou, huyiming2016, zhuzheng2014, huangjunjie2016\}@ia.ac.cn
}\\
{\tt\small \{liang.xu, guan.huang, dalong.du\}@horizon.ai
}
{\tt\small \{changmanyu\}@stu.xmu.edu.cn
}}
\maketitle
%\thispagestyle{empty}

%%%%%%%%% ABSTRACT
\begin{abstract}
Existing methods in video action recognition mostly do not distinguish human body from the environment and easily overfit the scenes and objects. In this work, we present a conceptually simple, general and high-performance framework for action recognition in trimmed videos, aiming at person-centric modeling. The method, called Action Machine, takes as inputs the videos cropped by person bounding boxes. It extends the Inflated 3D ConvNet (I3D) by adding a branch for human pose estimation and a 2D CNN for pose-based action recognition, being fast to train and test. Action Machine can benefit from the multi-task training of action recognition and pose estimation, the fusion of predictions from RGB images and poses. On NTU RGB-D, Action Machine achieves the state-of-the-art performance with top-1 accuracies of 97.2\% and 94.3\% on cross-view and cross-subject respectively. Action Machine also achieves competitive performance on another three smaller action recognition datasets: Northwestern UCLA Multiview Action3D, MSR Daily Activity3D and UTD-MHAD. Code will be made available.
\end{abstract}

%%%%%%%%% BODY TEXT
\section{Introduction}

With the release of Kinetics-400 \cite{kinetics17} and Kinetics-600 \cite{Kinetics600} in the last two years, action recognition in videos has shown similar trend as the object recognition due to the ImageNet~\cite{imagenet}. A variety of tasks including trimmed video classification \cite{NonLocal2018}, temporal action recognition in untrimmed videos~\cite{BSN}, spatial-temporal action detection \cite{AVA}, have been quite popular in recent competitions~\cite{activitynet17,activitynet18}.

\begin{figure}[t]
\centering
\subfigure[]{\includegraphics[width=0.24\linewidth]{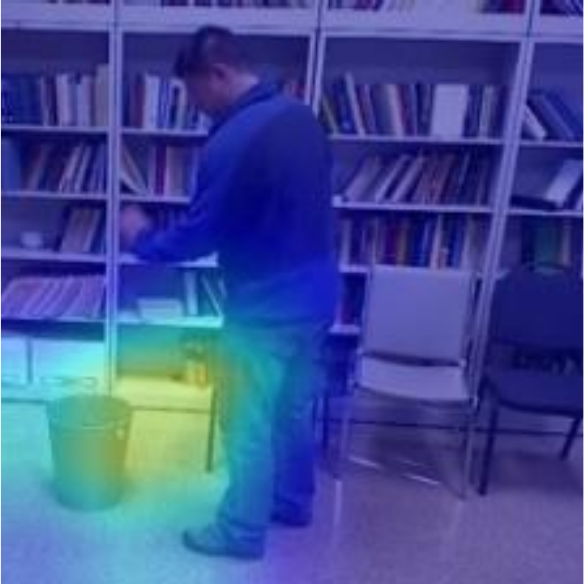}}
\subfigure[]{\includegraphics[width=0.24\linewidth]{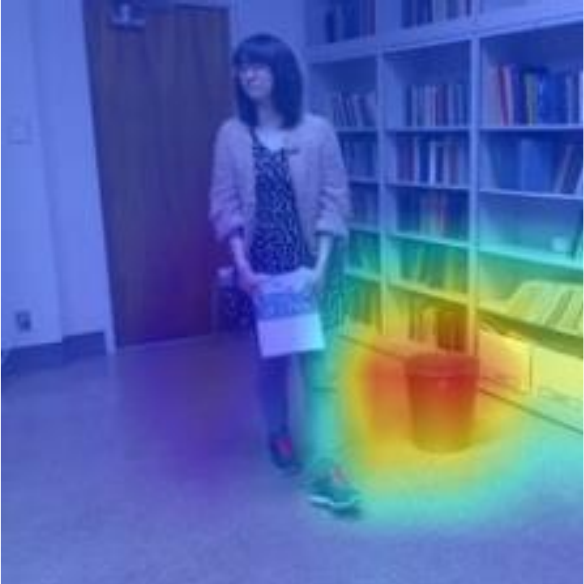}}
\subfigure[]{\includegraphics[width=0.24\linewidth]{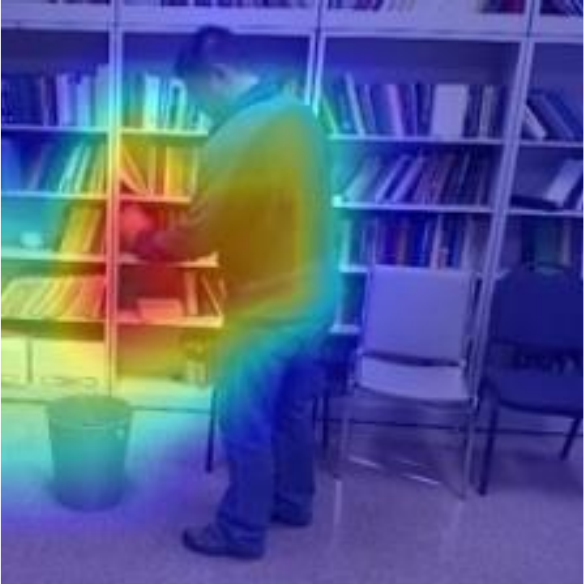}}
\subfigure[]{\includegraphics[width=0.24\linewidth]{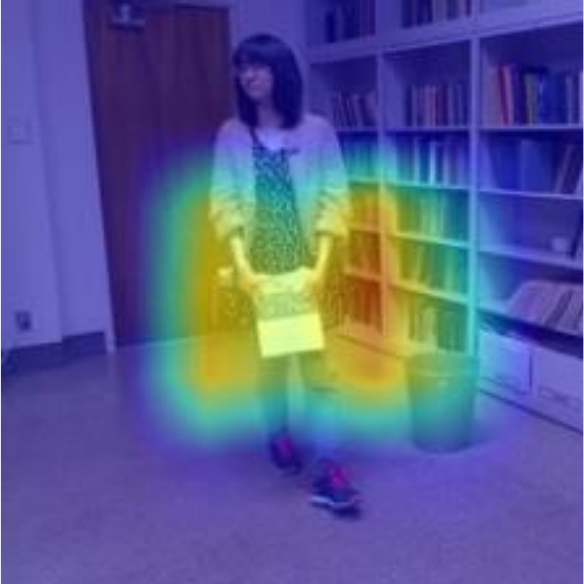}}
\caption{Visualizing the class-specific activation maps of Inflated 3D ConvNet (I3D)~\cite{kinetics17} with the Class Activation Mapping (CAM)~\cite{cam}. The video frames of two action classes from Northwestern UCLA Multiview Action3D~\cite{UCLAmultiview} are displayed, \ie, \textit{drop trash}, \textit{carry}, which are acted by a man and a woman respectively. The results of our person-centric modeling method (subfigure (c) and (d)) are more related to body movements, while the baseline I3D (subfigure (a) and (b)) overfits the \textit{trash can}.}
\label{Figure0}
\end{figure}

This paper studies action recognition in trimmed videos. To some extent, advances in this field are hampered by the biases in datasets collection, lack of annotations and object recognition in images~\cite{imagenet}. For example, the videos in UCF-101 \cite{UCF101} and HMDB-51 \cite{HMDB51} are rich in scenes and objects, while missing person bounding box annotations.\footnote{Except their subsets, UCF-24~\cite{UCF101} and JHMDB-21~\cite{jhmdb}, which are for spatial-temporal action detection.} Previous methods~\cite{NIPS2014_twostream,TSN2016ECCV,DTPP2018ICPR,Gated2018ICPR,kinetics17}, which do not directly distinguish human body from videos, tend to predict an action according to the scenes and objects, since convolutional neural networks (CNNs) make it easier to classify the objects
and things than human motions. They can be easily distracted by some irrelevant cues of videos when recognizing an action. As shown in Fig.~\ref{Figure0}(b), the video frame with ground-truth class \textit{carry} is predicted as a wrong action \textit{drop trash} by the baseline Inflated 3D ConvNet (I3D)~\cite{kinetics17}. Because the model has learned that the \textit{trash can} and the action \textit{drop trash} always appear in a video together (Fig.~\ref{Figure0}(a)).

This motivates us to design a model that can explicitly capture human body movements from videos, simultaneously follows the stream of RGB and CNN-based methods in action recognition. Pose (skeleton) data is lightweight, easy to understand and highly relevant to human action. It can be readily estimated by deep models, due to the recent advances of human pose estimation in a single image \cite{coco,hourglass,cpn,simple,2d3d,googlepose} and in videos \cite{simple}. The pose estimation methods are usually based on person bounding boxes, which can greatly filter out non-human clutters in RGB images. In view of this, person-centric action recognition has the potential to benefit from the joint training with pose estimation. Thanks to the large-scale annotated datasets~\cite{coco} and powerful deep networks~\cite{resnet}, the poses estimated from images in the wild are more robust than the skeleton captured by depth sensor like Kinect which is limited to indoor pose-based action recognition~\cite{trust16,viewadaptive,zhuco16,duyong15,viewnn18,stgcn}. Thus, action recognition in videos can be naturally formulated as a multi-task learning problem including RGB-based action recognition, pose estimation and pose-based action recognition.

In this work, a person-centric modeling method for human action recognition is proposed, called Action Machine, which shares similar spirit with Convolutional Pose Machines (CPM)~\cite{cpm} in sequential fashion model design. The proposed method (Fig.~\ref{Figure1}) extends the Inflated 3D ConvNet (I3D)~\cite{kinetics17} by adding a branch for human pose estimation and a 2D CNN for pose-based action recognition. In details, the video frames are cropped by the bounding boxes of target persons and are taken as the inputs of I3D. For frame-wise pose estimation, a 2D deconvolution head is added to the last convolutional layer of I3D, in parallel with the existing head for RGB-based action recognition. After the pose estimation task, a 2D CNN is applied to the pose sequences for pose-based action recognition. At test time, the predictions of two classification heads are fused by summation.
Some class-specific activation maps of Action Machine are shown in Fig.~\ref{Figure0}(c) and (d), indicating only the regions that really correspond to the action are activated. The main contributions of this work are summarized as follows:

\begin{enumerate}
	\item We present a conceptually simple and general framework for action recognition in trimmed videos, called Action Machine, aiming at person-centric modeling.
	\item The proposed techniques of explicitly modeling human body movements including person cropping, multi-task training of action recognition and pose estimation, the fusion of predictions from RGB images and poses can help to improve the model performance.
	\item We showcase the generality of our framework via extensive experiments on four human action datasets. Action Machine achieves the state-of-the-art performance on NTU RGB-D~\cite{ntu2016}, Northwestern UCLA Multiview Action3D~\cite{UCLAmultiview}. It also achieves competitive performance on MSR Daily Activity3D~\cite{MSRdaily} and UTD-MHAD~\cite{utd}. Action Machine is a high-performance framework while being fast to train and test.
\end{enumerate}

In the remainder of this paper, related works are given in Section~\ref{RelatedWorks}. Section~\ref{Methods} describes our proposed approach. In Section~\ref{Experiments}, our method is evaluated on the datasets. Finally, discussions and conclusions are given in Section~\ref{Conclusion}.

\begin{figure*}[t]
\centering
\includegraphics[width=0.8\linewidth]{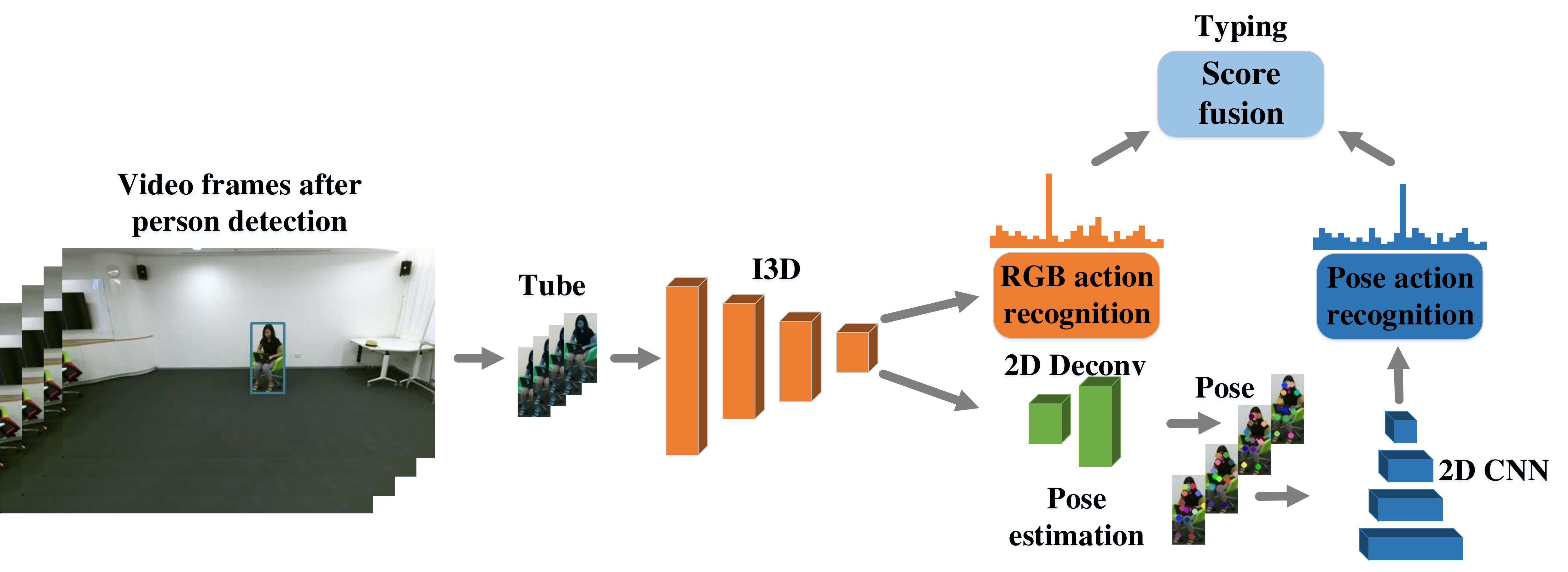}
\caption{Action Machine. It consists of the following steps: First, the videos after person cropping are used as the inputs of I3D for RGB-based action recognition. Then a 2D deconvolution head is added to the last convolutional layer of I3D for frame-wise pose estimation. Third, the estimated pose sequences are fed into a 2D CNN for pose-based action recognition. The proposed method is trained in a multi-task manner. Finally, the predictions of two heads for action recognition are fused by summation at test time.}
\label{Figure1}
\end{figure*}

%-------------------------------------------------------------------------
\section{Related works}
\label{RelatedWorks}

%\subsection{Hand-crafted features for action recognition}
%Before the surge of deep learning, the hand-crafted features are dominant in action recognition. They can be categorized into two categories: holistic representations \cite{MEI,STV,actionbank} and local representations~\cite{Laptev2005,cuboids,DT13,IDT13}. Holistic representations are the global representation of human body structure, shape and movements, such as Motion Energy Image (MEI) and Motion History Image (MHI)~\cite{MEI}, Space-Time Volume (STV)~\cite{STV}, action bank~\cite{actionbank}. Local representations is based on the extraction of local features, \eg, Space Time Interest Points ~\cite{Laptev2005}, Cuboids~\cite{cuboids}, Trajectories~\cite{DT13,IDT13}.

\subsection{Deep learning for action recognition}
{\bf RGB-based methods}. Two-stream ConvNet \cite{NIPS2014_twostream} employs RGB images and optical flow stacks as the inputs of two networks and fuses their predictions by late fusion. Temporal Segment 	 Network (TSN)~\cite{TSN2016ECCV} improves the performance of two-stream ConvNet by sparsely sampling video frames and learning video-level predictions. Deep networks with Temporal Pyramid Pooling (DTPP)~\cite{DTPP2018ICPR} samples enough frames from videos and learns video-level representation end-to-end. Using one network, C3D~\cite{C3D} learns spatial-temporal patterns from video clips by 3D convolutions. In 2017, DeepMind released a large-scale video action datasets Kinetics~\cite{kinetics17} and proposed Inflated 3D ConvNet (I3D). Non-local Net \cite{NonLocal2018} equips I3D with attention mechanism, extracting long-range interactions in spatial-temporal domain. The above models take as inputs the random spatial crops of video frames during training and can easily overfit the scenes and objects in videos because of failing to focus on human body explicitly. Different from them, we use the detected bounding boxes to crop the target persons from videos as the inputs of model, eliminating the effect of background context.

{\bf Pose (Skeleton)-based methods}. Compared with RGB images, skeleton data has the merits of being lightweight and free from scene cues.
Previous studies on pose-based action recognition can be categorized into RNN-based~\cite{trust16,viewadaptive,zhuco16}, CNN-based \cite{duyong15,viewnn18} and GCN-based(Graph Convolution Network)~\cite{stgcn} methods. RNN-based \cite{trust16,viewadaptive,zhuco16} methods treat the skeleton data as vectors and capture the sequence information of skeleton. CNN-based methods \cite{duyong15,anew2017} represent a skeleton sequence as a pseudo-image and recognize the underlying action in the same way as image classification. GCN-based methods \cite{stgcn} capture joint interactions on the skeleton graphs, explicitly considering the adjacent relationship among joints in a non-Euclidean space. In this work, we follow the CNN-based methods \cite{duyong15,viewnn18} and use the 2D CNN for the pose-based action recognition.

\subsection{Human pose estimation}
Human pose estimation can fall into top-down methods~\cite{cpn,simple,hourglass,he2017maskrcnn} in which a pose estimator is applied to the output of a person detector, and bottom-up methods~\cite{openpose,cpm}, in which keypoint proposals are grouped together into person instances. In this work, we adopt a top-down method and resort to a off-the-shelf detector~\cite{dai17dcn} for bounding boxes. For pose estimation, a 2D deconvolution head is added to the last convolutional layer of I3D. This is inspired from Mask R-CNN~\cite{he2017maskrcnn}, which extends Faster R-CNN~\cite{fasterrcnn} to support keypoint estimation. Action Machine does not involve detection task during training and the person cropping operations are imposed on the images instead of features.

\subsection{Multi-task learning for action recognition}
Chained multi-stream network \cite{chained} unifies three sources: RGB images, optical flow and body part mask for action recognition and detection. It introduces a Markov chain model to fuse these cues successively. In~\cite{2d3d}, Soft-argmax is extended to regress 2D and 3D pose directly, leading to the end-to-end training of pose estimation and action recognition. Different from the above two works, Action Machine is based on I3D, which has less parameters than C3D~\cite{C3D} because of 2D+1D convolution \cite{NonLocal2018}. It is also easy to train because of transferring pre-trained weights from 2D CNN and does not need the costly optical flow maps compared to two-stream ConvNet~\cite{NIPS2014_twostream}. The pose estimation method we use is detection-based, detecting keypoint by regressing heatmap and can get more accurate pose than the regression-based pose estimation in~\cite{2d3d}. Meanwhile, the pose estimation head can benefit from the temporal context of the I3D output.

\section{Action Machine}
\label{Methods}

\begin{figure*}[t]
\centering
\includegraphics[width=0.8\linewidth]{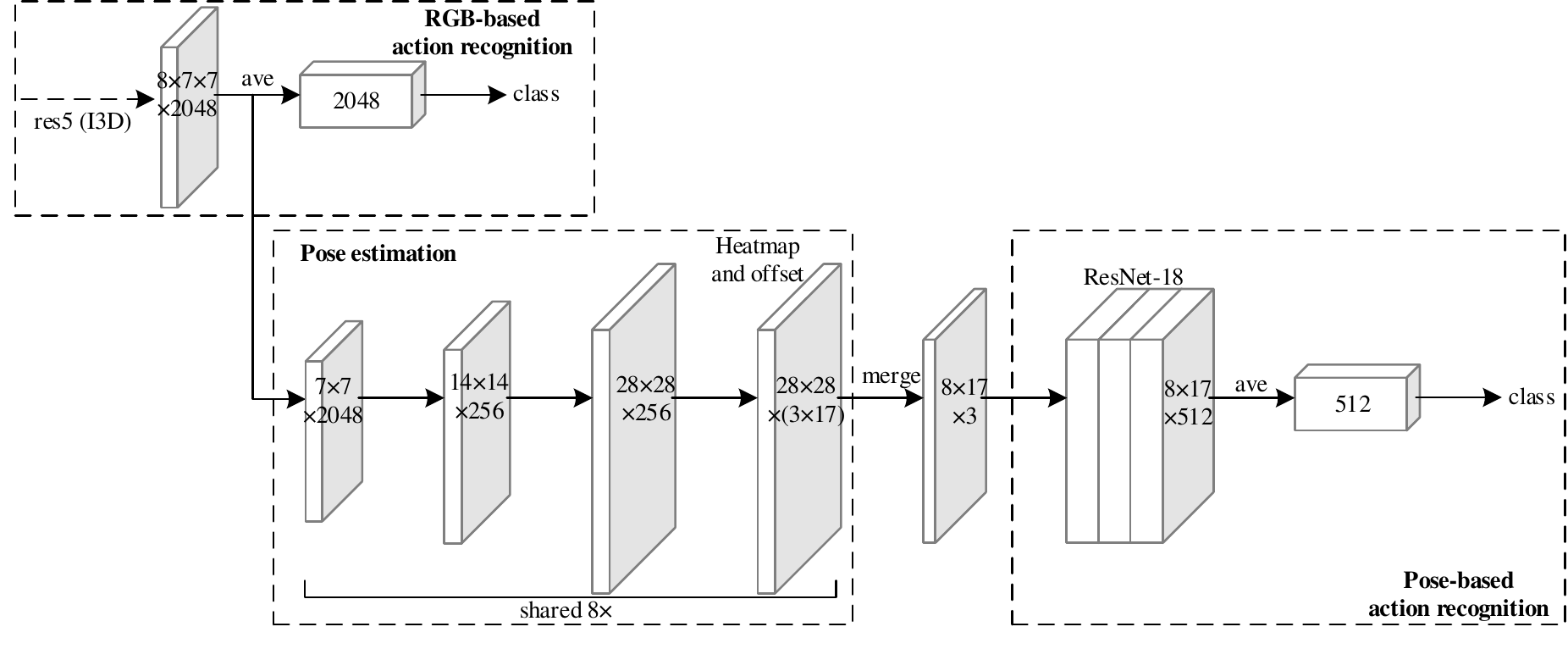}
\caption{We extend I3D by adding a branch for human pose estimation and a 2D CNN for pose-based action recognition. Numbers denote spatial resolution and channels. Arrows denote either conv, deconv, or fc layers as can be inferred from context (conv preserves spatial dimension while deconv increases it). The output conv of heatmap and offsetmap is 1$\times$1, deconvs are 4$\times$4 with stride 2. `res5' denotes the fifth stage of I3D with ResNet-50. `8$\times$' denotes the shared operations of 2D pose estimation on the temporal dimension.}
\label{Figure2}
\end{figure*}

As shown in Fig.~\ref{Figure1}, the pipeline of Action Machine consists of the following steps: First, the videos after person cropping are taken as the inputs of I3D for RGB-based action recognition. Then a 2D deconvolution head is added to the last convolutional layer of I3D for frame-wise pose estimation. The heatmaps produced by the pose estimation head are converted into joint coordinates by an argmax operation. Third, the transformed joint coordinates with 2D shape are taken as inputs by a 2D CNN for pose-based action recognition. The proposed method is trained in a multi-task manner. Finally, two sources of predictions, \ie, RGB images and poses, are fused by summation at test time.

{\bf Network input}. All the video frames are fed to a published detector, \ie, Deformable CNN~\cite{dai17dcn} for person bounding boxes and the category confidence threshold is set to 0.99 to avoid most false positive detections. The minimal bounding box enclosing all the detected boxes in a video is used for person cropping. In this way, the problem of detection-missing in videos is mostly solved. And cropping video by a shared box for all frames can align the feature on the temporal dimension.

{\bf Backbone}. We use the I3D with ResNet-50~\cite{resnet} backbone shown in Table~\ref{Table-1} for feature extraction. In order to estimate the pose of each frame, we remove the temporal max pooling after the first stage of I3D. The output feature map of the backbone has a size of 2048$\times$8$\times$7$\times$7, used both by RGB-based action recognition and pose estimation.

\begin{table}
\begin{center}
\scalebox{0.8}[0.8]{
\setlength{\tabcolsep}{1mm}{
\begin{tabular}{c|c|c}
    & layer & output size \\
\hline
${\rm conv_{1}}$ & 5${\times}$7${\times}$7, 64, stride 1, 2, 2 & 8${\times}$112${\times}$112 \\
\hline
${\rm pool_{1}}$ & 1${\times}$3${\times}$3 max, stride 1, 2, 2 & 8${\times}$56${\times}$56 \\
\hline
${\rm res_{2}}$  &
$\Bigg[
\begin{array}{c}
3{\times}1{\times}1, 64  \\
1{\times}3{\times}3, 64 \\
1{\times}1{\times}1, 256
\end{array}
\Bigg]{\times}3$            & 8${\times}$56${\times}$56 \\
\hline
${\rm res_{3}}$  & $\Bigg[
\begin{array}{c}
3{\times}1{\times}1, 128   \\
1{\times}3{\times}3, 128 \\
1{\times}1{\times}1, 512
\end{array}
\Bigg]{\times}4$             & 8${\times}$28${\times}$28 \\
\hline
${\rm res_{4}}$  & $\Bigg[
\begin{array}{c}
3{\times}1{\times}1, 256  \\
1{\times}3{\times}3, 256 \\
1{\times}1{\times}1, 1024
\end{array}
\Bigg]{\times}6$             & 8${\times}$14${\times}$14 \\
\hline
${\rm res_{5}}$  & $\Bigg[
\begin{array}{c}
3{\times}1{\times}1, 512  \\
1{\times}3{\times}3, 512 \\
1{\times}1{\times}1, 2048
\end{array}
\Bigg]{\times}3$             & 8${\times}$7${\times}$7 \\

\end{tabular}}}
\end{center}
\caption{Our used ResNet-50 I3D model for video. The dimensions
of 3D output maps and filter kernels are in T$\times$H$\times$W (2D
kernels in H$\times$W), with the number of channels following. The
input size is 8$\times$224$\times$224. Residual blocks are shown in brackets.}
\label{Table-1}
\end{table}

{\bf RGB-based action recognition}. As shown in Fig.~\ref{Figure2}, global average pooling is performed after the last convolutional layer of I3D to get a 2048-d feature $P_{rgb}$.

Consider a dataset of $N$ videos with $n$ categories $\{(X_i,y_i)\}_{i=1}^{N}$, where $y_i\in\{1,\ldots,n\}$ is the label. Formally, the prediction can be obtained directly

\begin{equation}\label{eq1}
  Y_{rgb}=\varphi(W_cP_{rgb}+b_c),
\end{equation}
where $\varphi$ is the softmax operation, ~$Y_{rgb}\in{\mathbb{R}^n}$. $W_c$ and $b_c$ are the parameters of the fully connected layer. In the training stage, combining with cross-entropy loss, the final loss function is

\begin{equation}\label{eq2}
  L_r=-\sum_{i=1}^{N}\log(Y_{rgb}(y_i)),
\end{equation}
where $Y_{rgb}(y_i)$ is the value of the $y_i$-th dimension of $Y_{rgb}$.

{\bf Pose estimation}. Given the output features of I3D, the pose estimation is performed on each temporal dimension. Inspired from Mask R-CNN~\cite{he2017maskrcnn}, a 2D deconvolution head is added to the last convolutional layer of I3D, as shown in Fig.~\ref{Figure2}. By default, two deconvolutional layers with batch normalization~\cite{BN} and ReLU activation \cite{AlexNet} are used. Each layer has 256 filters with 4$\times$4 kernel and the stride is 2. Following~\cite{googlepose}, a 1$\times$1 convolutional layer is added at last to generate predicted heatmaps for all $K$ keypoints (one channel per keypoint) and offsets (two channels per keypoint for the $x$ and $y$-directions) for a total of 3$K$ output channels, where $K = 17$ is the number of keypoints.

Given the image crop, let $f_k(x_i) = 1$ if
the $k$-th keypoint is located at position $x_i$ and 0 otherwise.
Here $k \in {1, . . . ,K}$ indexes the keypoint type and $i \in {1, . . . ,Q}$ indexes the pixel locations on the
image crop grid. For each position $x_i$ and each keypoint $k$, we compute the probability $h_k(x_i) = 1$ if $||x_i-l_k||\le M$, which means the
point $x_i$ is within a disk of radius $M$ from the location $l_k$ of
the $k$-th keypoint. $K$ such heatmaps are trained by solving a binary classification problem for each position and keypoint independently. For each position $x_i$ and each keypoint $k$, we also predict the 2D offset vector $F_k(x_i) = l_k - x_i$ from the pixel to the corresponding keypoint. $K$ such vector fields are trained by solving a 2D regression problem for each position and keypoint independently.

The output of the heatmap branch yields the heatmap probabilities $h_k(x_i)$ for each position $x_i$ and each keypoint $k$. The training target for the heatmap branch $\overline{h}_k(x_i)$ is a map of zeros and ones, with $\overline{h}_k(x_i) = 1$ if $||x_i - l_k|| \le M$ and 0 otherwise. The corresponding loss function $L_h(\theta)$ is the sum of smooth $L_1$ loss for each position and keypoint independently

\begin{equation}\label{eq4}
\begin{aligned}
  L_h(\theta) = \frac{1}{K}\sum_{k=1}^{K}\sum_{i}R(h_k(x_i),\overline{h}_k(x_i)),
\end{aligned}
\end{equation}
where $R$ is the smooth $L_{1}$ loss.

For training the offset regression branch, the differences between the predicted and ground truth offsets are penalized by smooth $L_1$ loss. The offset loss is only computed for positions $x_i$ within a disk of radius $M$ from each keypoint.

\begin{equation}\label{eq5}
\begin{aligned}
  L_o(\theta) =  \frac{1}{K}\sum_{k=1}^{K}\sum_{i:||l_k-x_i||{\le}M}R(F_k(x_i),(l_k-x_i)),
\end{aligned}
\end{equation}

The final loss function for pose estimation has the form

\begin{equation}\label{eq5}
\begin{aligned}
  L_p = {\lambda}_hL_h(\theta) + {\lambda}_oL_o(\theta),
\end{aligned}
\end{equation}
where ${\lambda}_h = 0.5$ and ${\lambda}_o = 0.5$ are two scalar factors to balance the loss function.

At test time, for the $k$-th keypoint, the argmax operation is performed on the $k$-th heatmap to yield the coarse location
\begin{equation}\label{eqargmax}
\begin{aligned}
x_k = \argmax{x_i}(h_k(x_i),i \in {1, . . . ,Q}).
\end{aligned}
\end{equation}

The accurate coordinate of the $k$-th keypoint is obtained by adding the corresponding offset $F_{k}(x_k)$ to $x_k$.

%\begin{equation}\label{eq5}
%\begin{aligned}
%x_k=x_k+F_{k}(x_k).
%\end{aligned}
%\end{equation}

{\bf Pose-based action recognition}. The coordinates of 2D pose can be transformed into a tensor of a size $2{\times}T{\times}K$, where $T$ denotes the number of input frames. An extra confidence channel is added for each predicted joints, which is obtained by max pooling over the heatmap and passed to the ReLU activation. Then the tensor is fed into the ResNet-18~\cite{resnet} for pose-based action recognition, as shown in Fig.~\ref{Figure2}. Due to the low spatial dimension of the input pose sequences, in the used ResNet-18, all the pooling operations are removed and all the stride 2 operations in the convolutional layers are replaced with 1. Global average pooling is performed after the last convolutional layer of ResNet-18 to get a 512-d feature. The prediction of pose stream $Y_{paction}$ is optimized with cross-entropy loss
%
%\begin{table}
%\begin{center}
%\scalebox{0.8}[0.8]{
%\setlength{\tabcolsep}{1mm}{
%\begin{tabular}{ccc}
%\hline
%    & layer & output size \\
%\hline\hline
%${\rm conv_{1}}$ & 7${\times}$7, 64, stride 1, 1 & 8${\times}$17 \\
%\hline\hline
%${\rm res_{2}}$  &
%$\Big[
%\begin{array}{c}
%3{\times}3, 64 \\
%3{\times}3, 64 \\
%\end{array}
%\Big]{\times}2$            & 8${\times}$17 \\
%\hline\hline
%${\rm res_{3}}$  & $\Big[
%\begin{array}{c}
%3{\times}3, 128 \\
%3{\times}3, 128 \\
%\end{array}
%\Big]{\times}2$             & 8${\times}$17 \\
%\hline\hline
%${\rm res_{4}}$  & $\Big[
%\begin{array}{c}
%3{\times}3, 256 \\
%3{\times}3, 256 \\
%\end{array}
%\Big]{\times}2$             & 8${\times}$17 \\
%\hline\hline
%${\rm res_{5}}$  & $\Big[
%\begin{array}{c}
%3{\times}3, 512 \\
%3{\times}3, 512 \\
%\end{array}
%\Big]{\times}2$             & 8${\times}$17 \\
%
%\hline
%\end{tabular}}}
%\end{center}
%\caption{Our used ResNet-18 C2D model for pose action recognition. The dimensions
%of 2D output maps and filter kernels are in H$\times$W, with the number of channels following. The
%input size is 8$\times$17.}
%\label{Table-2}
%\end{table}

\begin{equation}\label{eq6}
  L_{paction}=-\sum_{i=1}^{N}\log(Y_{paction}(y_i)).
\end{equation}

{\bf Multi-task training}. Action Machine has three tasks: RGB-based action recognition, pose estimation and pose-based action recognition. They are jointly optimized by the following loss function:

\begin{equation}\label{eq7}
\begin{aligned}
  L = \lambda_1L_r + \lambda_2L_p + \lambda_3L_{paction},
\end{aligned}
\end{equation}
where $\lambda_1$, $\lambda_2$ and $\lambda_3$  are the loss weights of RGB-based action recognition, pose estimation and pose-based action recognition respectively.\footnote{Note that the gradients of pose-based action recognition don't backpropagate into the pose estimation head because of the argmax operation in equation~\ref{eqargmax}. We do not use the Soft-argmax in~\cite{2d3d} on heatmap for end-to-end training because we find the keypoints quality of this approach on COCO~\cite{coco} is lower than ours.} They are all set to 1.0 by default.

{\bf Fusion of RGB and pose-based action recognition}. In order to combine the strengths of predictions from RGB images and poses, the predicted probabilities of two heads are fused by summation at test time.

%\begin{equation}\label{eq8}
%\begin{aligned}
%  Y = Y_{rgb} + Y_{paction}.
%\end{aligned}
%\end{equation}

\section{Experiments}
\label{Experiments}
%-------------------------------------------------------------------------
\subsection{Datasets}

The proposed method has been evaluated on five datasets: on COCO~\cite{coco} for pose estimation, and on four trimmed video action datasets: NTU RGB-D~\cite{ntu2016}, Northwestern-UCLA Multiview Action 3D~\cite{UCLAmultiview}, MSR Daily Activity3D~\cite{MSRdaily} and UTD-MHAD~\cite{utd} for action recognition.\footnote {Because the videos in these datasets usually have one person (except the 11 classes of NTU RGB-D where the videos have two persons), which are appropriate for validating the effect of person-centric modeling only by simple human detection. Datasets like UCF-101 \cite{UCF101} and HMDB-51 \cite{HMDB51} have some group activities (sports) with more than one person and a variety of scenes. It may need tracking and person re-identification for our use and additional annotation cost, which is beyond the scope of this paper.}

{\bf COCO}~\cite{coco}. The COCO train, validation, and test sets contain more than 200k images and 250k person instances annotated with
keypoints. 150k instances of them are publicly available for training and validation. Our models are trained on COCO train2017 dataset (includes 57K
images and 150K person instances) and tested on the val2017 set.

{\bf NTU RGB-D}~\cite{ntu2016}. This dataset is acquired with a Kinect v2 sensor. It contains more than 56K videos and 4 million frames with 60 different activities including individual activities, interactions between two people, and health-related events. The actions are performed by 40 subjects and recorded from 80 viewpoints. We follow the cross-subject and cross-view protocol from~\cite{ntu2016}.

{\bf Northwestern-UCLA Multiview Action 3D (N-UCLA)}~\cite{UCLAmultiview}. This dataset contains 1494 sequences, covering 10 action categories, such as drop trash or sit down. Each sequence is captured simultaneously by 3 Kinect v1 cameras. Each action is performed one to six times by ten different subjects. We follow the cross-view protocol defined by~\cite{UCLAmultiview}. It has three cross-view combinations: \emph{xview1}, \emph{xview2} and \emph{xview3}. The combination \emph{xview1} means that the samples from view 2 and 3 are for training, and the samples from view 1 are for testing.

{\bf MSR Daily Activity3D (MSR)}~\cite{MSRdaily}. This dataset contains 320 videos shot with a Kinect v1 sensor. 16 daily activities are performed twice by 10 subjects from a single viewpoint. Following~\cite{MSRdaily}, we use the videos from subject 1, 3, 5, 7 and 9 for training, and the remaining ones for testing.

{\bf UTD-MHAD}~\cite{utd}. This dataset is collected using a Microsoft
Kinect sensor and a wearable inertial sensor in an
indoor environment. It contains 27 actions performed by 8
subjects and has 861 sequences. Cross-subject protocol~\cite{utd} is used for testing.

\subsection{Experimental settings for pose estimation}

{\bf Training.} The ground truth human box is made to a fixed aspect ratio (height : width = 4 : 3) by extending the box in height or width. It is then
cropped from the image and resized to a fixed resolution. The default resolution is
384$\times$288. Data augmentation includes scale($\pm$30\%), rotation($\pm$30 degrees) and flip. Our models are pre-trained on ImageNet \cite{imagenet}. ResNet-50 is used by default. We train our models on a 4-GPU machine and each GPU has 2 images in a mini-batch (so in total with a mini-batch size of 8 images). We train our models for 22 epochs in total, starting with a learning rate of 0.01 and reducing it by a factor of 10 according to a schedule of [17, 21]. SGD is used, with a momentum of 0.9 and a weight decay of 0.0001. The L1 norm of all gradients is clipped by 2 independently. MXNet~\cite{mxnet} is used for implementation.

{\bf Testing.} The detected person bounding boxes on COCO val2017 are used. Following the common practice in~\cite{hourglass,cpn,simple}, the joint location is predicted on the averaged heatmaps of the original images and their horizontal flips. Following~\cite{cocoleaderboard}, we use the mean average precision (AP) over 10 OKS (object keypoint similarity) thresholds for evaluation.
\subsection{Experimental settings for action recognition}

{\bf Preprocessing}. As shown in Table~\ref{Table-3}, for all datasets, we resize their videos to make them smaller and keep their aspect ratios. The sampling stride is selected according to the frame rate of videos and model performance.

{\bf Training}. Our models are pre-trained on ImageNet \cite{imagenet}. Then the pre-trained weights are inflated from 2D to 3D, as shown in Table \ref{Table-1}. For small datasets, including N-UCLA~\cite{UCLAmultiview}, MSR~\cite{MSRdaily} and UTD-MHAD~\cite{utd}, we also try pre-training our models on NTU RGB-D~\cite{ntu2016}. The models are fine-tuned using 8-frame clips with sampling stride shown in Table~\ref{Table-3}. The start frame is randomly sampled during training. Data augmentation includes the bounding box center, width, height jittering and random mirror. Then the bounding box is made to a fixed aspect ratio (height : width = 1 : 1\footnote {We do not keep the same aspect ratio as 2D pose estimation in COCO, because we use a shared box for all frames in a video and a moving person in a video is likely to cover a range different from a still person in a single image.}) by extending the box in height or width. It is then cropped from the image and resized to a fixed resolution. The default resolution is 224$\times$224. We train our models on a 4-GPU machine and each GPU has 4 clips (32 images) in a mini-batch (in total with a mini-batch size of 16 clips). We train our models for 85 epochs in total, starting with a learning rate of 0.01 and reducing it by a factor of 10 according to a schedule of [42, 68]. SGD is used, with a momentum of 0.9 and a weight decay of 0.0001. The L1 norm of all gradients is clipped by 2 independently. Dropout with ratio of 0.5 is used before the fully connected layer of RGB-based and pose-based action recognition. The pose annotations of video frames are obtained by using the detection boxes and models trained on COCO~\cite{coco}. MXNet~\cite{mxnet} is used for implementation.

{\bf Testing}. Following~\cite{NonLocal2018}, the fully convolutional inference is performed spatially on videos, including three crops, \ie, the up left, center, down right of bounding box center. 10 clips are evenly sampled from a full-length video and the softmax scores are computed on them individually. The final prediction is the averaged softmax scores of all clips. We report the top-1 accuracy using the model of the last epoch.

\begin{table}
\begin{center}
\scalebox{0.8}[0.8]{
\setlength{\tabcolsep}{1mm}{
\begin{tabular}{ccccc}
\hline
 Dataset   & resolution & resize to & frame rate & stride\\
\hline\hline
NTU RGB-D~\cite{ntu2016} & 1080$\times$1920    & 256$\times$454 & 30 & 8 \\
N-UCLA~\cite{UCLAmultiview}    & 480$\times$640      & 256$\times$340 & 12 & 1 \\
MSR~\cite{MSRdaily}       & 480$\times$640      & 256$\times$340 & 30 & 8 \\
UTD-MHAD~\cite{utd}  & 480$\times$640      & 256$\times$340 & 15 & 4 \\
\hline
\end{tabular}}}
\end{center}
\caption{Video preprocessing and configurations.}
\label{Table-3}
\end{table}

\subsection{Pose estimation on COCO}
\begin{figure}[t]
\centering
\includegraphics[width=1\linewidth]{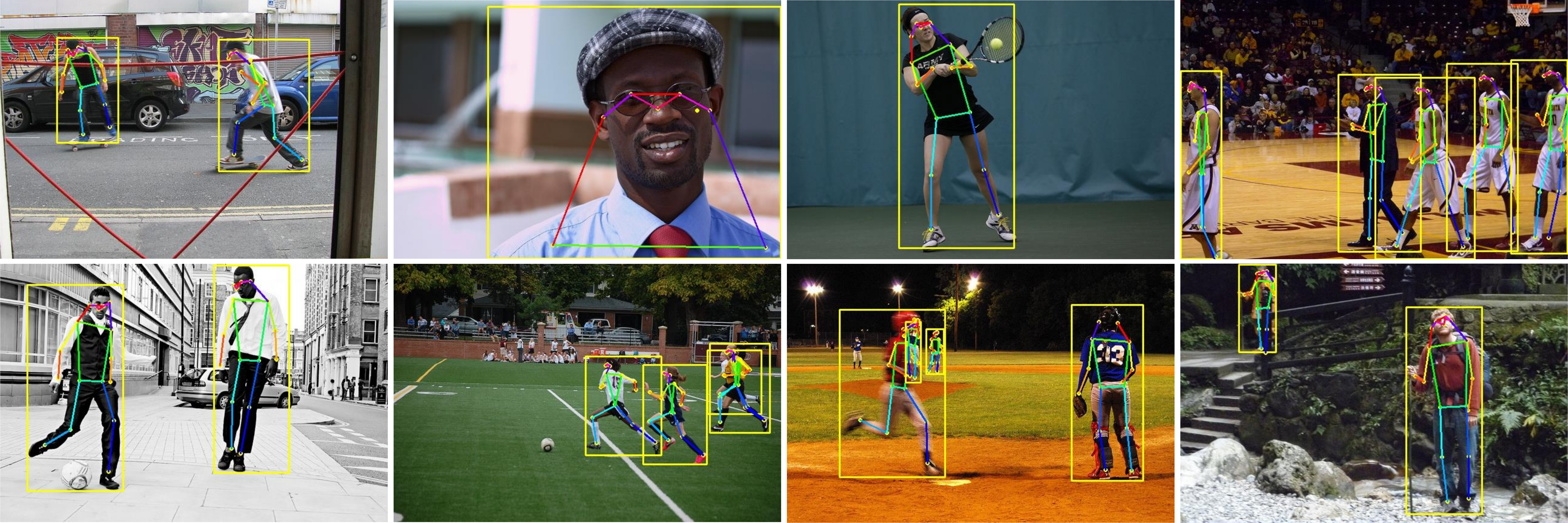}
\caption{Results of our pose estimation method on COCO~\cite{coco}.}
\label{Figurecoco}
\end{figure}

\begin{table}
\begin{center}
\scalebox{0.8}[0.8]{
\setlength{\tabcolsep}{1mm}{
\begin{tabular}{cccc}
\hline
Method             & Backbone  & Input Size     & AP \\
\hline
8-stage Hourglass~\cite{hourglass}  & -         & 256$\times$192 & 66.9 \\
%CPN~\cite{cpn}                & ResNet-50 & 256$\times$192 & 68.6 \\
CPN~\cite{cpn}                & ResNet-50 & 384$\times$288 & 70.6 \\
%Simple Baseline~\cite{simple}    & ResNet-50 & 256$\times$192 & 70.6 \\
Simple Baseline~\cite{simple}    & ResNet-50 & 384$\times$288 & 72.2 \\
\hline
%Ours               & ResNet-50 & 256$\times$192 & 68.8  \\    %67.6
Ours               & ResNet-50 & 384$\times$288 & 72.7  \\    %71.5
\hline
\end{tabular}}}
\end{center}
\caption{Comparison with Hourglass~\cite{hourglass}, CPN~\cite{cpn} and Simple Baseline~\cite{simple} on COCO val2017 dataset.
%Their results are cited from~\cite{simple}.
}
\label{Table-10}
\end{table}

As shown in Table~\ref{Table-10}, our method is compared with state-of-the-art methods: Hourglass~\cite{hourglass}, CPN~\cite{cpn} and Simple Baseline~\cite{simple} on COCO val2017.
Our method achieves competitive performance with the above methods. Fig.~\ref{Figurecoco} shows some results of our pose estimation method on COCO val2017 dataset. The proposed method equipped with I3D is used for the next action recognition experiments.

\subsection{Action recognition}

In this section, Action Machine is compared with other approaches on four human action datasets.
Results are shown in Table~\ref{Table-4}, \ref{Table-5}, \ref{Table-6}, \ref{Table-utd}, where $\checkmark$ denotes that the corresponding modality is used as the input of model in \textit{testing}. Note that Action Machine does not take as input human poses in testing because it has learned to estimate poses from RGB images after training.

{\bf Performance on NTU RGB-D.} In Table \ref{Table-4}, our model is compared with pose-based methods \cite{Lie14,hcn15,ntu2016,ntu2016,trust16,tcn17,anew2017,viewadaptive,stgcn,srtsl} and RGB-based methods~\cite{dssca,chained,2d3d,glimpse,posemap}. Action Machine with single modality (RGB) as input at test time achieves the state-of-the-art performance. Specifically, Action Machine outperforms PoseMap \cite{posemap} by 2 and 2.6 points in top-1 accuracy on cross-view and cross-subject respectively. Compared to PoseMap \cite{posemap}, Action Machine is conceptually simple and easy to implement.

\begin{table}
\begin{center}
\scalebox{0.8}[0.8]{
\setlength{\tabcolsep}{1mm}{
\begin{tabular}{ccccc}
\hline
    &Pose & RGB & xview & xsub \\
\hline
Lie Group~\cite{Lie14}& $\checkmark$  & - &	52.8 &	50.1\\
H-RNN~\cite{hcn15} & $\checkmark$ & - &	64.0 &	59.1\\
Deep LSTM~\cite{ntu2016}& $\checkmark$ & - &	67.3 &	60.7\\
PA-LSTM~\cite{ntu2016} & $\checkmark$ & - &	70.3 &	62.9\\
ST-LSTM+TS~\cite{trust16}& $\checkmark$ & - &	77.7 &	69.2\\
Temporal Conv~\cite{tcn17}& $\checkmark$ & - &	83.1 &	74.3\\
C-CNN+MTLN~\cite{anew2017}& $\checkmark$ & - &	84.8 &	79.6\\
VA-LSTM~\cite{viewadaptive} &$\checkmark$ & - &	87.6 &	79.4 \\
ST-GCN~\cite{stgcn} &$\checkmark$ & - &	88.3 &	81.5 \\
SR-TSL~\cite{srtsl} &$\checkmark$ & - &	92.4 &	84.8 \\
%Hands Attention~\cite{handsattention}& $\checkmark$& $\checkmark$ & 90.6 & 84.8 \\
\hline
Chained~\cite{chained}& $\checkmark$ & - &	- &	80.8\\
DSSCA-SSLM~\cite{dssca}& $\checkmark$ & $\checkmark$ &	- &	74.9\\
2D-3D-Softargmax~\cite{2d3d}& - & $\checkmark$ &   - & 85.5    \\
Glimpse Clouds~\cite{glimpse}& - & $\checkmark$ &   93.2 & 86.6    \\
PoseMap~\cite{posemap}& $\checkmark$ & $\checkmark$ &   95.2 & 91.7    \\
\hline\hline
{\bf Action Machine (Ours)}& - & $\checkmark$ & {\bf 97.2}    & {\bf 94.3}  \\
\hline
\end{tabular}}}
\end{center}
\caption{Performance on NTU RGB-D, accuracy({\%}).}
\label{Table-4}
\end{table}

{\bf Performance on N-UCLA.} In Table \ref{Table-5}, our model is compared with pose-based methods: \cite{Lie14,hcn15,enhancedviz,ensembledeep} and RGB-based methods~\cite{glimpse}. Action Machine with single modality (RGB) as input outperforms previous state-of-the-art approaches. Without using LSTM and extra handcrafted rules as Glimpse Clouds \cite{glimpse}, Action Machine has a accuracy gain of 4.7 points in average top-1 accuracy on cross-view.

\begin{table}
\begin{center}
\scalebox{0.8}[0.8]{
\setlength{\tabcolsep}{1mm}
{
\begin{tabular}{ccccccc}
\hline
 & Pose & RGB & xview1 & xview2 & xview3 & Avg \\
\hline
Lie Group~\cite{Lie14} &  $\checkmark$ & -  &    -    &   -    &  -   & 74.2 \\
H-RNN~\cite{hcn15}     &  $\checkmark$ & - &    -    &   -    &  -   & 78.5 \\
Enhanced viz.~\cite{enhancedviz} & $\checkmark$ & - &    -    &    -   &  -   & 86.1 \\
Ensemble TS-LSTM~\cite{ensembledeep}& $\checkmark$ & - &  -    &   -    &  -   & 89.2 \\
\hline
Glimpse Clouds~\cite{glimpse} & - & $\checkmark$ & 83.4 & 89.5 & 90.1 & 87.6      \\
\hline\hline
{\bf Action Machine (Ours)} &  -  & $\checkmark$ & {\bf 88.3} & {\bf 92.2} & {\bf 96.5} & {\bf 92.3} \\
\hline
\end{tabular}}}
\end{center}
\caption{Performance on N-UCLA, accuracy({\%}).}
\label{Table-5}
\end{table}

{\bf Performance on MSR.} In Table \ref{Table-6}, our model is compared with pose-based methods \cite{MSRdaily,efficientposebased,movingpose,movingposelet}, depth-based methods \cite{depthfusion,mmmp,dlgsgc,dssca}. Without using depth modality, Action Machine achieves competitive performance compared to DSSCA-SSLM \cite{dssca}, which is based on handcrafted feature including RGB and depth. However, on the cross-subject of NTU RGB-D (Table~\ref{Table-4}), a larger dataset than MSR, DSSCA-SSLM \cite{dssca} is lower than ours for 19.4 points, showing the robustness of our method against the amount of data.

\begin{table}
\begin{center}
\scalebox{0.8}[0.8]{
\setlength{\tabcolsep}{1mm}{
\begin{tabular}{ccccc}
\hline
  & Pose & RGB & Depth & xsub \\
\hline
Action Ensemble~\cite{MSRdaily} & $\checkmark$ & - & - &68.0 \\
Efficient Pose-Based~\cite{efficientposebased} & $\checkmark$ & - & - &73.1 \\
Moving Pose~\cite{movingpose}       & $\checkmark$ & - & - &73.8 \\
Moving Poselets~\cite{movingposelet} & $\checkmark$ & - & - & 74.5 \\
\hline
Depth Fusion~\cite{depthfusion}  & - & - & $\checkmark$& 88.8 \\
MMMP~\cite{mmmp}              & $\checkmark$ & -  & $\checkmark$ &91.3 \\
DL-GSGC~\cite{dlgsgc}        & $\checkmark$ & - & $\checkmark$&95.0 \\
DSSCA-SSLM~\cite{dssca}     & - & $\checkmark$     & $\checkmark$ &{\bf 97.5} \\
%\hline
%Hands Attention~\cite{handsattention}& $\checkmark$& $\checkmark$ & - & 90.6 \\
\hline\hline
{\bf Action Machine (Ours)}  & - &  $\checkmark$ & - &93.0  \\
\hline
\end{tabular}}}
\end{center}
\caption{Performance on MSR Daily Activity3D, accuracy({\%}).}
\label{Table-6}
\end{table}

{\bf Performance on UTD-MHAD.} In Table \ref{Table-utd}, our model is compared with~\cite{jtm,OpticalSpectra,jdm,posemap}. Without using the 3D pose extracted by depth sensor as these methods, Action Machine with RGB modality as input achieves competitive performance.

\begin{table}
\begin{center}
\scalebox{0.8}[0.8]{
\setlength{\tabcolsep}{1mm}{
\begin{tabular}{ccccc}
\hline
  & Pose & RGB & xsub \\
\hline
JTM~\cite{jtm} & $\checkmark$ & - &85.8 \\
Optical Spectra~\cite{OpticalSpectra} & $\checkmark$ & - &86.9 \\
JDM~\cite{jdm} & $\checkmark$ & - & 88.1 \\
PoseMap~\cite{posemap}& $\checkmark$ & $\checkmark$ & {\bf 94.5}    \\
\hline\hline
{\bf Action Machine (Ours)}  & - & $\checkmark$ &92.5  \\
\hline
\end{tabular}}}
\end{center}
\caption{Performance on UTD-MHAD, accuracy({\%}).}
\label{Table-utd}
\end{table}

\subsection{Ablation study}

Ablation studies are performed on NTU RGB-D and N-UCLA to verify the effectiveness of our techniques for person-centric modeling in Action Machine. There are four basic configurations, as illustrated below:

\begin{figure*}[t]
\centering
\subfigure[Activation maps of {\bf RGBAction person crop}]{\includegraphics[width=0.49\linewidth]{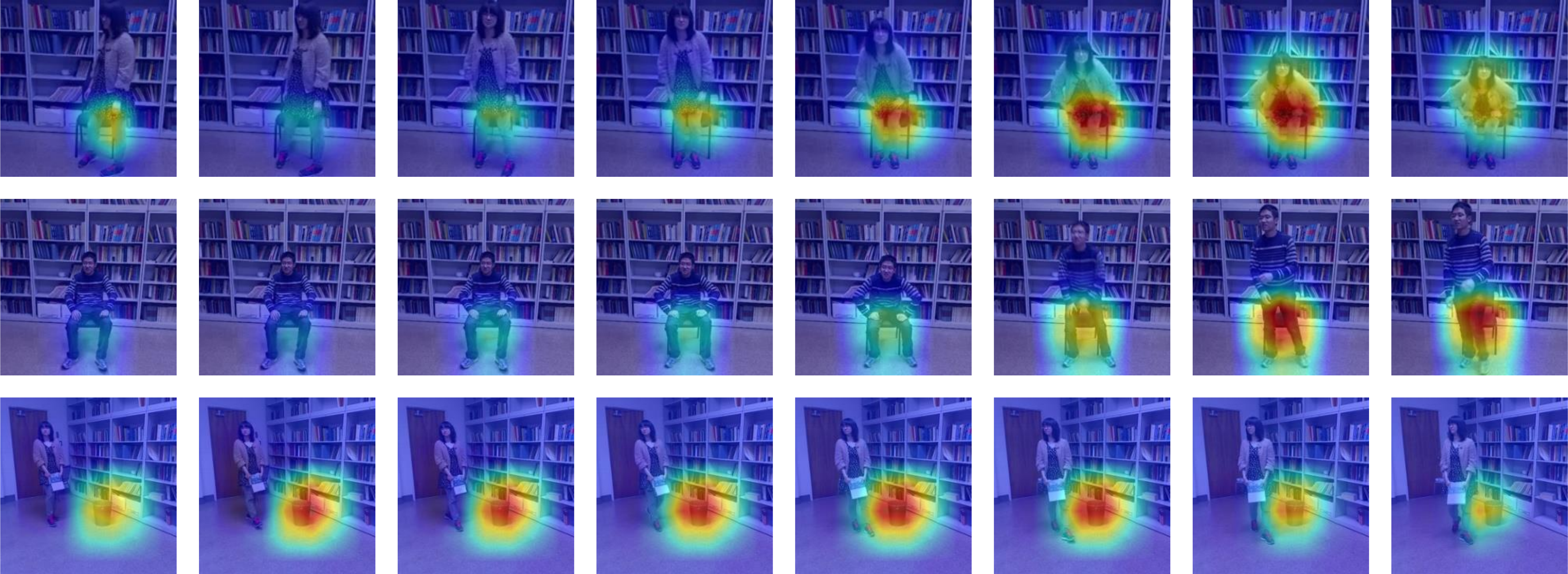}}
~
\subfigure[Activation maps of {\bf KPS RGBAction}]{\includegraphics[width=0.49\linewidth]{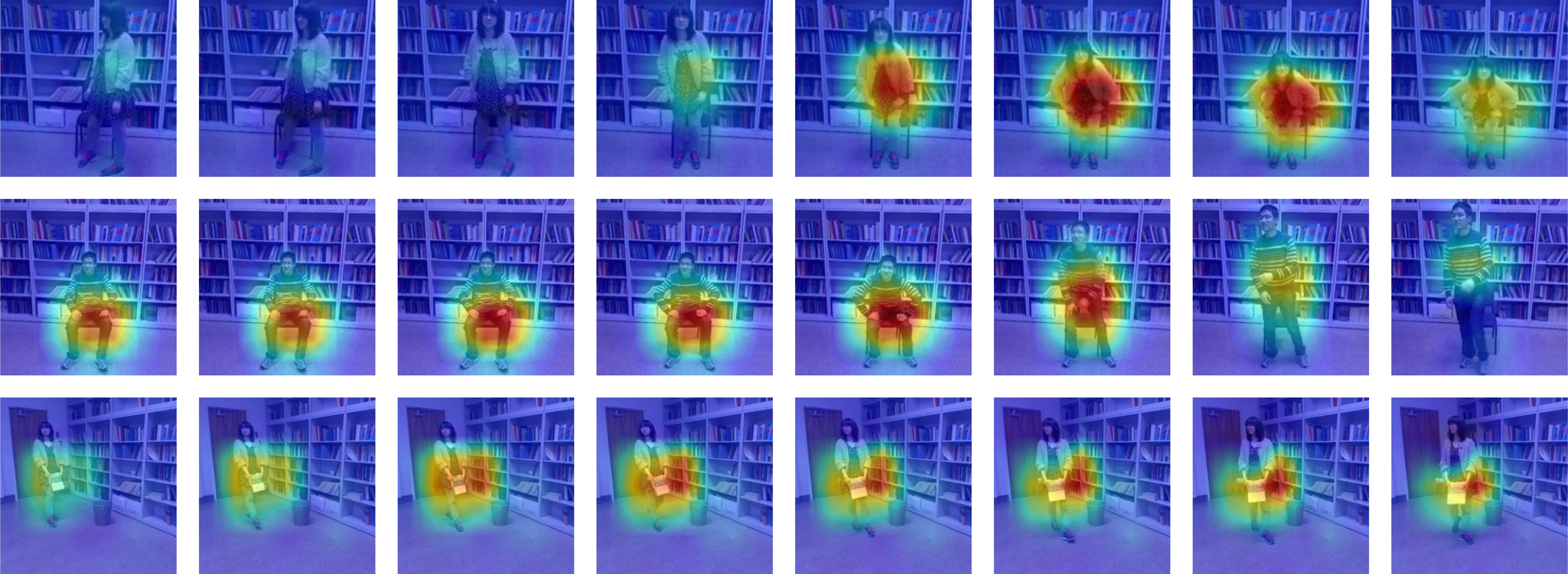}}
\caption{Visualizing the class-specific activation maps of our model with the Class Activation Mapping (CAM)~\cite{cam}. Activation maps of video snippets of three action classes, \ie, \textit{sit down}, \textit{stand up}, \textit{carry} are shown from top to the bottom. It is clear that the multi-task training of RGB-based action recognition and pose estimation can make the model focus on the spatial-temporal regions related to the action class.}
\label{Figure3}
\end{figure*}

{\bf RGBAction random crop}. The baseline I3D model takes as inputs the random crops of videos and performs action recognition using RGB feature.

{\bf RGBAction person crop}. The I3D model takes as inputs the videos after person cropping and performs action recognition using RGB feature.

{\bf KPS RGBAction}. The I3D model takes as inputs the videos after person cropping, performs action recognition using RGB feature, and adds a head for pose estimation.

{\bf KPS PoseAction RGBAction}. The I3D model takes as inputs the videos after person cropping, adds a head for pose estimation, and performs action recognition using RGB and pose feature. The model trained from {\bf KPS RGBAction} is used as the pre-trained model. We fix it and only train the ResNet-18 or ResNet-50 for pose-based action recognition. We report the results of pose-based action recognition and the sum fusion of predictions from RGB images and poses.

As shown in Table~\ref{Table-7}, on the cross-subject of NTU RGB-D, person cropping can improve the model accuracy by 0.9 points over random crop.
Our full model outperforms the baseline {\bf RGBAction random crop} by 2 points. Due to the high accuracy of baseline, the improvement on the cross-view is not obvious. Similar gain potential can also be observed on the small subsets of NTU RGB-D (xview-s, xsub-s), which are originally used for the fast training and testing in our implementation.

\begin{table}
\begin{center}
\scalebox{0.8}[0.8]{
\setlength{\tabcolsep}{1mm}{
\begin{tabular}{c|cccc}
    & xview & xsub & xview-s & xsub-s \\
\hline
{\bf RGBAction random crop} & 97.2           & 92.3 & 94.3 & 61.2  \\
\hline
{\bf RGBAction person crop} & {\bf 97.7}    & 93.2 & 94.5 & 67.9  \\
\hline
{\bf KPS RGBAction}         & 97.3          & 93.8         & 95.0 & 71.2  \\
\hline
\tabincell{c}{{\bf KPS PoseAction RGBAction}\\ (ResNet-18)} & 90.1/97.1    & 84.9/94.1 & 87.8/95.9 & 62.9/72.7 \\
\hline
\tabincell{c}{{\bf KPS PoseAction RGBAction}\\ (ResNet-50)} & 91.3/97.2    & 85.5/{\bf 94.3}       & 89.9/{\bf 96.1} &  66.0/{\bf 73.5}\\
\end{tabular}}}
\end{center}
\caption{Ablation studies on NTU RGB-D, accuracy({\%}). In the rows which have slash $/$, the number on the left of slash is the accuracy of pose-based action recognition, the right is the accuracy of fusion of RGB and pose results. \emph{xview-s} and \emph{xsub-s} denote the small subsets of NTU RGB-D cross-view and cross-subject respectively.}
\label{Table-7}
\end{table}

As shown in Table~\ref{Table-8}, on N-UCLA, Action Machine outperforms the baseline by a large margin. Specifically, {\bf RGBAction person crop} with the person cropping technique can improve the accuracy by 1.6 and 4.3 points on \emph{xview1} and \emph{xview3} over the baseline {\bf RGBAction random crop} respectively. Person cropping does not bring accuracy gain on \emph{xview2}, because the test crops of front view images (Fig.~\ref{Figure3}, the first and second row) on this dataset are close to that cropped by person boxes. Jointly training pose estimation and RGB-based action recognition, \ie, {\bf KPS RGBAction}, can improve about 3 to 7 points. Overall, using ResNet-18, our final model exceeds the baseline by 7.2 points. By using a stronger backbone, \ie, ResNet-50 for pose-based action recognition and NTU RGB-D pre-training, the accuracies of our models, either solely by poses or the fusion of RGB images and poses, are further improved.

To better understand how our approach learns discriminative feature for action recognition, the class-specific activation maps of our models are visualized with the Class Activation Mapping (CAM)~\cite{cam} approach in Fig.~\ref{Figure3}. The videos are sampled from N-UCLA, including three classes (\textit{sit down}, \textit{stand up}, \textit{carry}). These maps show that, jointly training RGB-based action recognition with pose estimation can make the model focus on the motions of human body. For example, {\bf KPS RGBAction} (Fig.~\ref{Figure3}(b)) pays more attention on the standing and sitting process, while {\bf RGBAction person crop} (Fig.~\ref{Figure3}(a)) seems to focus on the object (chair). Particularly, in the \textit{carry} example, {\bf KPS RGBAction} is significantly activated only by the hand (center of the body). Nevertheless, {\bf RGBAction person crop} is activated by the trash can, leading to a wrong prediction (\textit{drop trash}).

\begin{table}
\begin{center}
\scalebox{0.8}[0.8]{
\setlength{\tabcolsep}{1mm}
{
\begin{tabular}{c|cccc}
%\hline
    & xview1 & xview2 & xview3 & Avg \\
\hline
{\bf RGBAction random crop}    & 81.6      & 82.4     &	86.3      & 83.4 \\
\hline
{\bf RGBAction person crop}    & 83.2      & 82.4     &	90.6      & 85.4 \\
\hline
{\bf KPS RGBAction}            & 86.3      & 90       &	94.9      & 90.4 \\
\hline
\tabincell{c}{{\bf KPS PoseAction RGBAction}\\ (ResNet-18)} & 79.7/87.5 & 81/90.4  &	87.5/94.1 & 82.7/90.6\\
\hline
\tabincell{c}{{\bf KPS PoseAction RGBAction}\\ (ResNet-50)} & 84.2/{\bf 89.6} & 81.8/90  &	88.4/94.3 & 84.8/91.3\\
\hline
\tabincell{c}{{\bf KPS PoseAction RGBAction}\\ (ResNet-18, NTU pre-training)}  & 85.5/88.6 &  88.0/91.6 & 93.2/{\bf 96.5} & 88.9/92.2 \\
\hline
\tabincell{c}{{\bf KPS PoseAction RGBAction}\\ (ResNet-50, NTU pre-training)}  & 83.8/88.3 &  87.6/{\bf 92.2} & 93.2/{\bf 96.5} & 88.2/{\bf 92.3} \\
%\hline
\end{tabular}}}
\end{center}
\caption{Ablation studies on N-UCLA, accuracy({\%}). In the rows which have slash $/$, the number on the left of slash is the accuracy of pose-based action recognition, the right is the accuracy of fusion of RGB and pose results.}
\label{Table-8}
\end{table}

\subsection{Timing}

{\bf Inference:} We train a ResNet-50-I3D model that shares features between RGB-based action recognition and pose estimation with two deconvolutional layers. And it is followed by a ResNet-50 for pose-based action recognition. This model runs at $\sim$55ms per clip (8 frames) on an Nvidia TitanX GPU. As the dimension of pose sequences is small, substituting ResNet-50 with ResNet-18 for pose-based action recognition don't cause much difference: it finally takes $\sim$50ms. I3D takes $\sim$30ms. Action Machine is fast to test and adds only a small overhead to I3D.

{\bf Training:} Action Machine is also fast to train. Training with ResNet-50-I3D on the cross-view of NTU RGB-D takes 32 hours (0.66s per 16 clips (128 frames) mini-batch) in our synchronized 4-GPU implementation.

\section{Discussions and Conclusions}
\label{Conclusion}

We propose a person-centric modeling method: Action Machine, for human action recognition in trimmed videos. It has three complementary tasks: RGB-based action recognition, pose estimation and pose-based action recognition. By using person bounding boxes and human poses, Action Machine achieves competitive performance compared with other approaches on four video action datasets~\cite{ntu2016,UCLAmultiview,MSRdaily,utd}. However, in our implementation, it is hard to discard non-human clutters strictly (\eg, the trash can in Fig.~\ref{Figure3}) because of the bounding box quality and other postprocessing steps. Besides, in our multi-task training, the ground-truth pose annotations are estimated by the model trained on COCO~\cite{coco} and may not be abundant enough for the training of pose estimation task due to the paucity of videos. The joint training of pose estimation on COCO and action recognition on videos may relieve the problem, as we can exploit the data richness of COCO.

%Additionally, future works also include extending Action Machine into spatial-temporal action detection for multi-person action recognition and online action recognition.

{\small
\bibliographystyle{ieee}
\bibliography{cited}
}

\centerline{\large{\bf{ Appendix}}}

\section{NTU RGB-D small subset setting}
In Section 4.6, for ablation studies of different configurations of our models, we use the small subsets of NTU RGB-D (\textit{xview-s}, \textit{xsub-s}), which are designed by us for the fast training and testing.

For \textit{xview-s}, the sample videos of the original cross-view split with subject ID larger than 5 are discarded. For this evaluation, the training and testing sets have
3, 839 and 1, 917 samples (about 1/10th of the full cross-view split), respectively.

For \textit{xsub-s}, based on the original cross-subject split, we pick all the samples of camera 1 and discard samples of cameras 2 and 3. The sample videos with subject ID larger than 10 are discarded.
For this evaluation, the training and testing sets have
4, 317 and 1, 439 samples (about 1/10th of the full cross-subject split), respectively.

\section{Cross-dataset recognition task}

In order to show the advantage of person-centric modeling over baseline, we further test our trained models on another different datasets. Specifically, we train our models on NTU RGB-D cross-subject and test them on the test sets of N-UCLA, MSR Daily and UTD-MHAD respectively. The shared category mappings between the smaller dataset and NTU RGB-D are shown in Table~\ref{Table-1},~\ref{Table-2},~\ref{Table-3} and the test videos are limited to have these ground-truth classes. Because of the different sources of videos, the scene contexts and objects in training dataset are largely different from the testing dataset. A model without capturing human body motion will behave worse than the model which learns to focus on. Results are shown in Table~\ref{Table-4}. It is clearly observed that our proposed person-centric modeling techniques including: person cropping, multi-task training of action recognition and pose estimation, the fusion of predictions from RGB images and poses can help to improve the performance of baseline model {\bf RGBAction random crop} on different datasets. Existing methods based on RGB images easily overfit the scenes and objects of specific datasets without focusing on human body movements, though they may have high performance. In contrast, Action Machine is more generalizable and extendable.

\begin{table}
\begin{center}
\scalebox{0.8}[0.8]{
\setlength{\tabcolsep}{1mm}
{
\begin{tabular}{c|c}
N-UCLA & NTU RGB-D \\
\hline
pick up with one hand & pickup   \\
pick up with two hands & pickup   \\
stand up & standing up (from sitting position)  \\
sit down & sitting down  \\
throw  & throw\\
\end{tabular}}}
\end{center}
\caption{Shared category mapping between N-UCLA and NTU RGB-D.}
\label{Table-1}
\end{table}

\begin{table}
\begin{center}
\scalebox{0.8}[0.8]{
\setlength{\tabcolsep}{1mm}
{
\begin{tabular}{c|c}
MSR Daily & NTU RGB-D \\
\hline
drink & drink water  \\
eat & eat meal/snack \\
read book & reading        \\
call cellphone & make a phone call/answer phone  \\
write on a paper & writing       \\
cheer up& cheer up       \\
stand up & standing up (from sitting position) \\
sit down & sitting down \\
\end{tabular}}}
\end{center}
\caption{Shared category mapping between MSR Daily and NTU RGB-D.}
\label{Table-2}
\end{table}

\begin{table}
\begin{center}
\scalebox{0.8}[0.8]{
\setlength{\tabcolsep}{1mm}
{
\begin{tabular}{c|c}
UTD-MHAD & NTU RGB-D \\
\hline
sit to stand & standing up (from sitting position) \\
stand to sit & sitting down   \\
\end{tabular}}}
\end{center}
\caption{Shared category mapping between UTD-MHAD and NTU RGB-D.}
\label{Table-3}
\end{table}

\begin{table}
\begin{center}
\scalebox{0.7}[0.7]{
\setlength{\tabcolsep}{1mm}
{
\begin{tabular}{c|ccc}
%\hline
    & N-UCLA & MSR Daily & UTD-MHAD\\
\hline
{\bf RGBAction random crop}  &	70.0 & 70.8  & \bf{100}   \\
\hline
{\bf RGBAction person crop}   &	70.0 & 78.4 & \bf{100}    \\
\hline
{\bf KPS RGBAction}          &	76.4 & 79.7  & \bf{100}    \\
\hline
\tabincell{c}{{\bf KPS PoseAction RGBAction}\\ (ResNet-18)} & 68.8/76.4  &	58.2/79.7  & \bf{100}/\bf{100}\\
\hline
\tabincell{c}{{\bf KPS PoseAction RGBAction}\\ (ResNet-50)} & 69.2/\bf{77.3}  &	63.2/\bf{81.0} & \bf{100}/\bf{100} \\
%\hline
\end{tabular}}}
\end{center}
\caption{Cross-dataset testing on N-UCLA (\textit{xview3}), MSR Daily and UTD-MHAD, accuracy({\%}). The models on the first column are trained on NTU RGB-D cross-subject.
In the rows which have slash $/$, the number on the left of slash is the accuracy of pose-based action recognition, the right is the accuracy of fusion of RGB and pose results.}
\label{Table-4}
\end{table}

\newpage

\section{Pose estimation results on action recognition datasets}
We visualize the video frames with detected boxes and estimated poses on action recognition datasets in Fig.~\ref{Figure1}, \ref{Figure2}, \ref{Figure3}, \ref{Figure4}. We use the testing model {\bf KPS RGBAction}, which performs RGB-based action recognition and pose estimation simultaneously. In general, the estimated poses are accurate.
\begin{figure*}[!htp]
\centering
\includegraphics[width=1.0\linewidth]{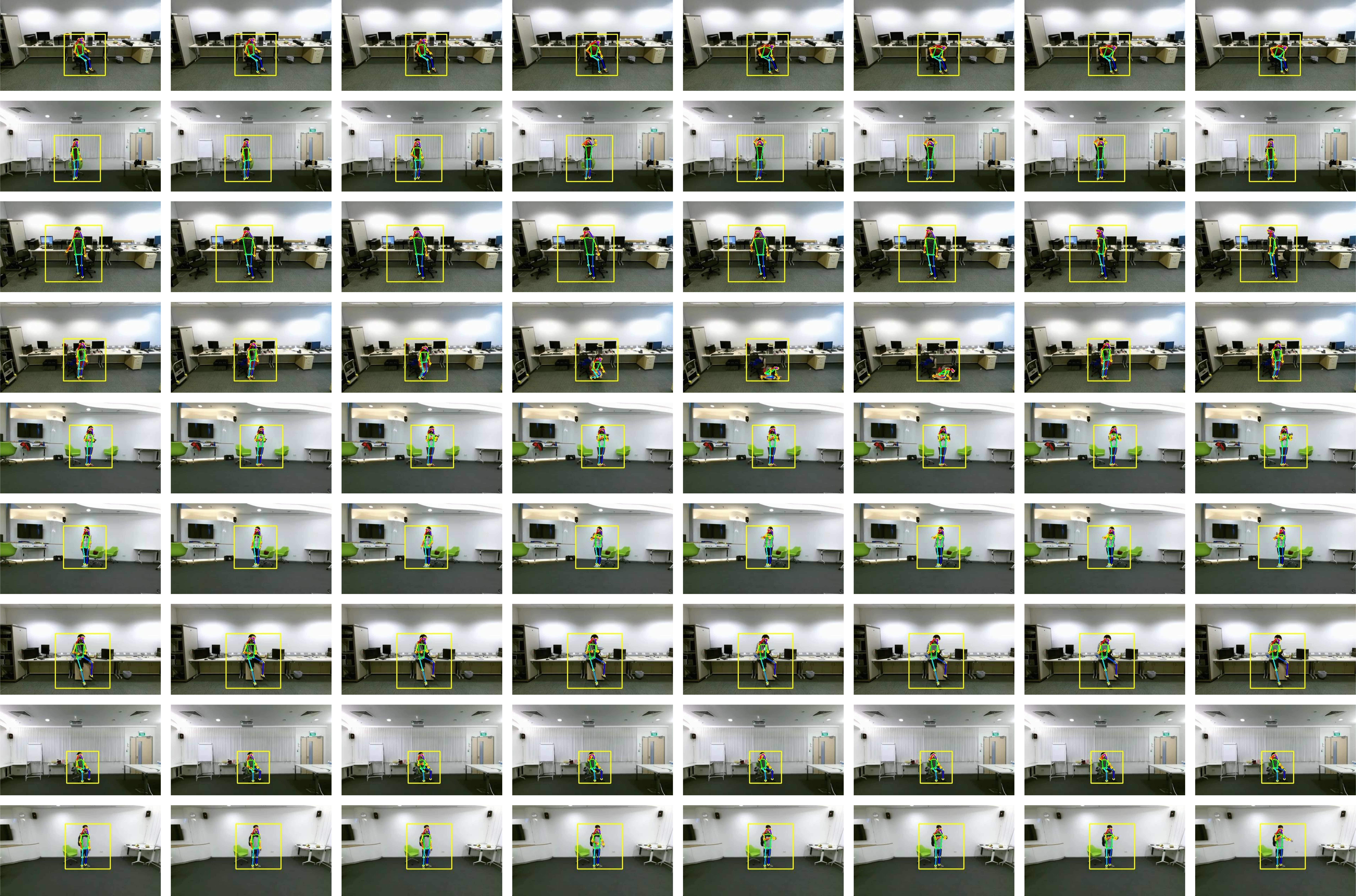}
\caption{Visualizing the video frames of the test set of NTU RGB-D cross-view. The video frames with detected boxes (yellow) and estimated poses of eight action classes, \ie, \textit{wear a shoe}, \textit{cheer up}, \textit{pointing to something with finger}, \textit{falling}, \textit{writing}, \textit{put the palms together}, \textit{reading}, \textit{brushing teeth}, \textit{tear up paper} are shown from top row to the bottom.}
\label{Figure1}
\end{figure*}

\begin{figure*}[!htp]
\centering
\includegraphics[width=1.0\linewidth]{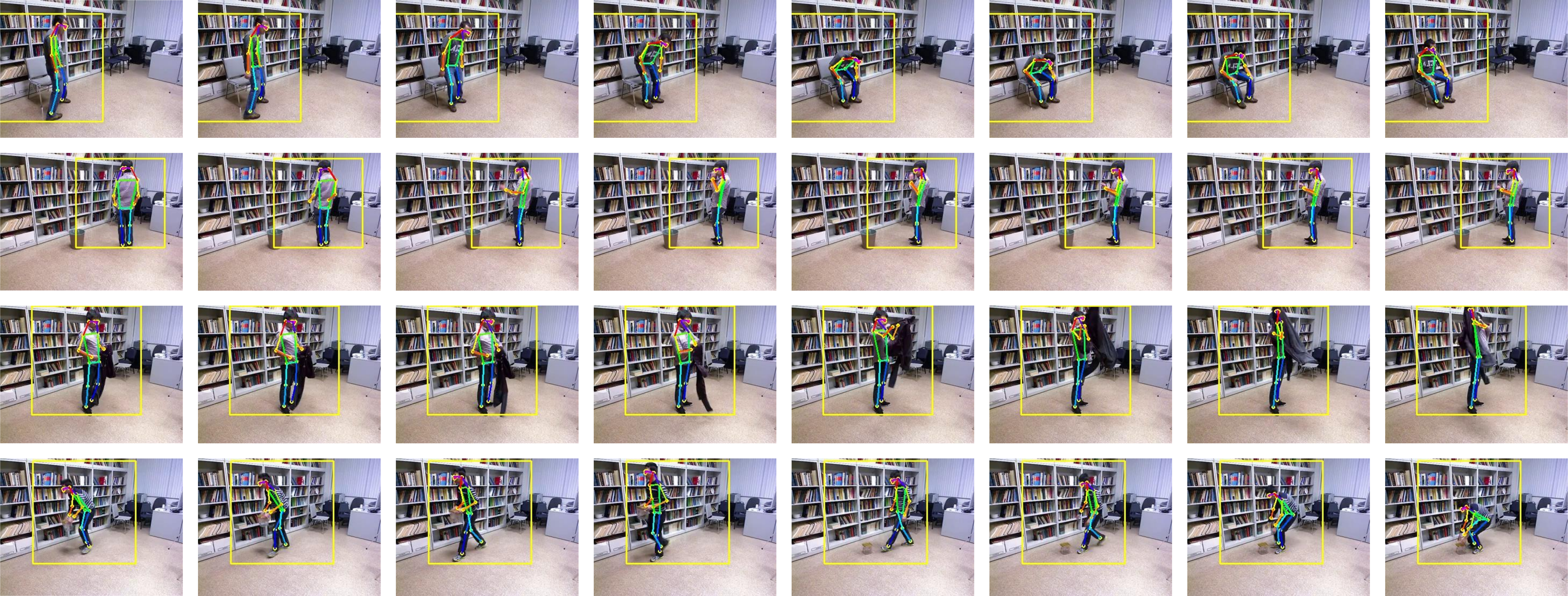}
\caption{Visualizing the video frames of the test set of N-UCLA \textit{xview3}. The video frames with detected boxes (yellow) and estimated poses of four action classes, \ie, \textit{sit down}, \textit{throw}, \textit{donning}, \textit{pick up with two hands} are shown from top row to the bottom.}
\label{Figure2}
\end{figure*}

\begin{figure*}[!htp]
\centering
\includegraphics[width=1.0\linewidth]{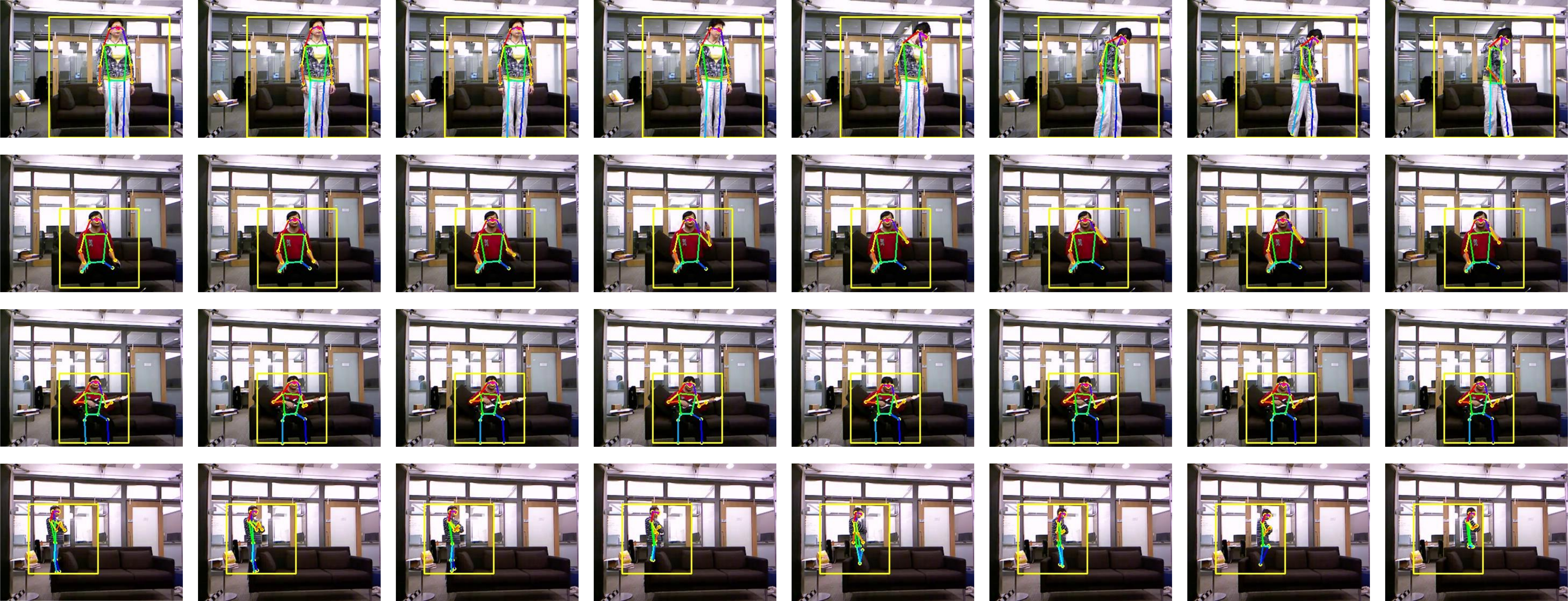}
\caption{Visualizing the video frames of the test set of MSR Daily. The video frames with detected boxes (yellow) and estimated poses of four action classes, \ie, \textit{sit down}, \textit{call cellphone}, \textit{play guitar}, \textit{walk} are shown from top row to the bottom.}
\label{Figure3}
\end{figure*}

\begin{figure*}[!htp]
\centering
\includegraphics[width=1.0\linewidth]{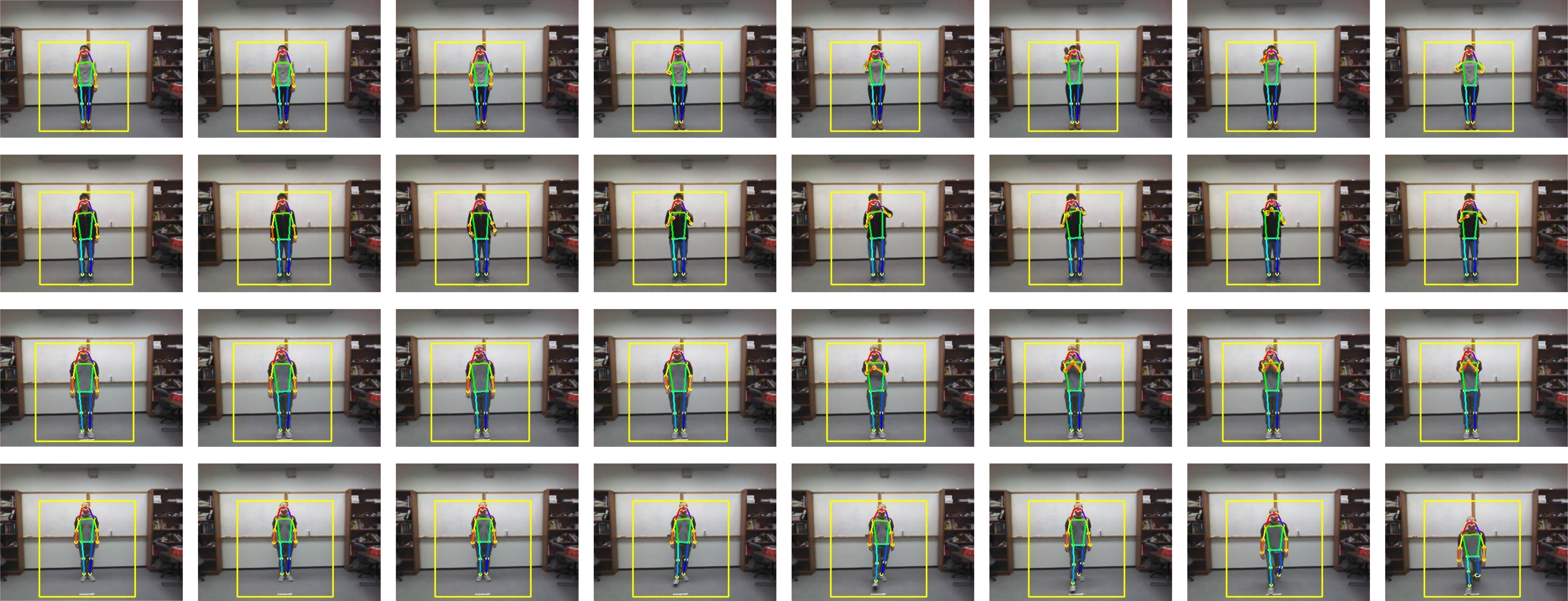}
\caption{Visualizing the video frames of the test set of UTD-MHAD. The video frames with detected boxes (yellow) and estimated poses of four action classes, \ie, \textit{two hand push}, \textit{front boxing}, \textit{cross arms in the chest}, \textit{forward lunge (left foot forward)} are shown from top row to the bottom.}
\label{Figure4}
\end{figure*}

\end{document}